\documentclass{article}

\usepackage{arxiv}

\usepackage[utf8]{inputenc} 
\usepackage[T1]{fontenc}    
\usepackage{hyperref}       
\usepackage{url}            
\usepackage{booktabs}       
\usepackage{amsfonts}       
\usepackage{nicefrac}       
\usepackage{microtype}      
\usepackage{lipsum}		
\usepackage{graphicx}
\usepackage{natbib}
\usepackage{doi}

\usepackage{siunitx}
\usepackage{multirow}
\usepackage{listings}
\usepackage{amsmath}
\usepackage{graphicx}
\usepackage{latexsym}
\usepackage{bmpsize}
\usepackage{url}
\usepackage{comment}
\usepackage{amsfonts}
\usepackage{bm}
\usepackage{subfig}
\usepackage{cite}
\usepackage{breqn}
\allowdisplaybreaks[1]

\usepackage{color}

\usepackage{xspace}

\def\erfc#1{{\rm erfc}({#1})}

\newcommand{\etal}{\textit{et al}.}

\title{Combining SNN and ANN for Enhanced Image Classification}

\author{
    Naoya Muramatsu \\
	Graduate School of Library,\\
    Information and Media Studies,\\
    University of Tsukuba\\
    Kasuga 1-2, Tsukuba-shi, Ibaraki, Japan\\
	\texttt{sh.mn.nat@gmail.com} \\
	\And
	Hai-Tao Yu \\
	Graduate School of Library,\\
    Information and Media Studies,\\
    University of Tsukuba\\
    Kasuga 1-2, Tsukuba-shi, Ibaraki, Japan\\
	\texttt{yuhaitao@slis.tsukuba.ac.jp} \\
}

\renewcommand{\shorttitle}{Combining SNN and ANN for Enhanced Image Classification}

\hypersetup{
pdftitle={A template for the arxiv style},
pdfsubject={q-bio.NC, q-bio.QM},
pdfauthor={David S.~Hippocampus, Elias D.~Striatum},
pdfkeywords={First keyword, Second keyword, More},
}

\begin{document}
\maketitle

\begin{abstract}
    With the continued innovations of deep neural networks, spiking neural networks (SNNs) that more closely resemble biological brain synapses have attracted attention owing to their low power consumption.
    However, for continuous data values, they must employ a coding process to convert the values to spike trains.
    Thus, they have not yet exceeded the performance of artificial neural networks (ANNs), which handle such values directly.
    To this end, we combine an ANN and an SNN to build versatile hybrid neural networks (HNNs) that improve the concerned performance.
    To qualify this performance, MNIST and CIFAR-10 image datasets are used for various classification tasks in which the training and coding methods changes.
    In addition, we present simultaneous and separate methods to train the artificial and spiking layers, considering the coding methods of each.
    We find that increasing the number of artificial layers at the expense of spiking layers improves the HNN performance.
    For straightforward datasets such as MNIST, it is easy to achieve the same performance as ANNs by using duplicate coding and separate learning.
    However, for more complex tasks, the use of Gaussian coding and simultaneous learning is found to improve the HNN’s accuracy while utilizing a smaller number of artificial layers.
\end{abstract}


\section{Introduction}

    Over the years, deep-learning methods have provided dramatic performance improvements for various tasks (e.g., image recognition~\citep{heDeepResidualLearning2016,heDelvingDeepRectifiers2015}, natural language processing~\citep{huConvolutionalNeuralNetwork2014,youngRecentTrendsDeep2018}, and perfect-information gaming~\citep{silverMasteringGameGo2016}).
    In fact, their performance has surpassed humans with certain tasks~\citep{silverMasteringGameGo2016,heDelvingDeepRectifiers2015}.
    However, deep learning continues to face challenges in terms of energy efficiency.
    For example, in recent machine-learning models, graphical processing units (e.g., NVIDIA Tesla V100 and A100) have been used to solve single tasks while consuming $\SI{250}{W}$.
    The human brain requires only ~$\SI{20}{W}$~\citep{drubach2000brain} for the same task while also managing related and routine activities.
    For this reason, spiking neural networks (SNNs), which represent information between neurons as spikes that more closely resemble the synaptic activity of biological neurons, have attracted attention.

    However, SNNs are inferior to artificial neural networks (ANNs) in terms of performance.
    There are several reasons for this.
    First, back-propagation cannot be directly applied to SNNs, because the spike generation mechanism is not differentiable.
    The main workaround is to use surrogate gradients.
    A few studies have shown performance improvements comparable to ANNs for limited tasks~\citep{NEURIPS2018_82f2b308,Wu2018,neftci2019surrogate}.
    In addition, we applied such a method to optimize weights using state-of-the-art approximated back-propagation~\citep{fang2020exploiting} for SNNs.
    The second reason for the SNN inferiority is that most of the data handled by SNNs consist of continuous values, such as those represented by floating point numbers.
    Furthermore, SNNs represent and process information via the synaptic spikes.
    This type of input/output would be ideal for process information from sensors that output similar spike signals, such as event-based cameras~\citep{Orchard2015,Gallego_2020}.
    Such a task would require an SNN-oriented visual image dataset, such as that of N-MNIST~\citep{Orchard2015}, so that higher accuracy and lower computation could be attained~\citep{Deng2020}.
    In contrast, many datasets (e.g., MNIST and CIFAR-10) and, for that matter, data handled in the real world often have normalized continuous values in $(0, 1)$ .
    Therefore, for SNNs to process the most realistic data, coding is required to convert the continuous values into spike trains.
    Many methods for doing this have been proposed, but their performance remains inferior to the ANNs, which process the original data directly~\citep{Deng2020}.

    As a novel method to address this problem, we propose the conception of hybrid neural networks (HNNs) in which an SNN are combined with an ANN.
    We use the ANN parts for the input and hidden layers and the SNN parts for the middle and output layers.
    In this way, the network can directly receive continuous-valued data via the ANN and compute them via the SNN, which is expected to realize a lower-energy and a higher-accuracy neural network.
    However, this structure of the network still requires the aforementioned coding process between the artificial layers (ALs) and the spiking layers (SLs), owing to the continuous values treated in the ALs and the spike signals in the SLs.
    The HNNs are evaluated using image classification tasks to determine whether ALs can be trained simultaneously with SLs.
    Furthermore, the ratio of ALs to SLs on a network and the coding methods needed between the ALs and SLs are evaluated.
    The results show that using more ALs improves the classification accuracy, and that the most straightforward coding method (i.e., duplicate coding) is effective in terms of network performance and computational cost.

    The following summarizes the major contributions of this work:
    \begin{enumerate}
        \item Because SNNs deal with continuous-valued data, the proposed HNN combines an ANN and an SNN; their ALs and SLs are connected via coding methods and can be trained separately or simultaneously via back-propagation.
        \item We show that increasing the ratio of ALs in the network improves HNN performance, which can surpass that of pure SNNs.
        \item We make a technical contribution by proposing the use of differentiable Gaussian coding. HNNs that use this method can achieve higher accuracy, even when the percentage of ALs is small.
    \end{enumerate}

\section{Background}

    In this section, we explain the theory behind SNNs while focusing on their learning algorithms and their method of informational representation.

    \subsection{Spiking Neural Networks}

    \subsubsection{Neural Structure}

\begin{figure}[htbp]
        \centering
        \subfloat[][]{\includegraphics[width=0.8\linewidth]{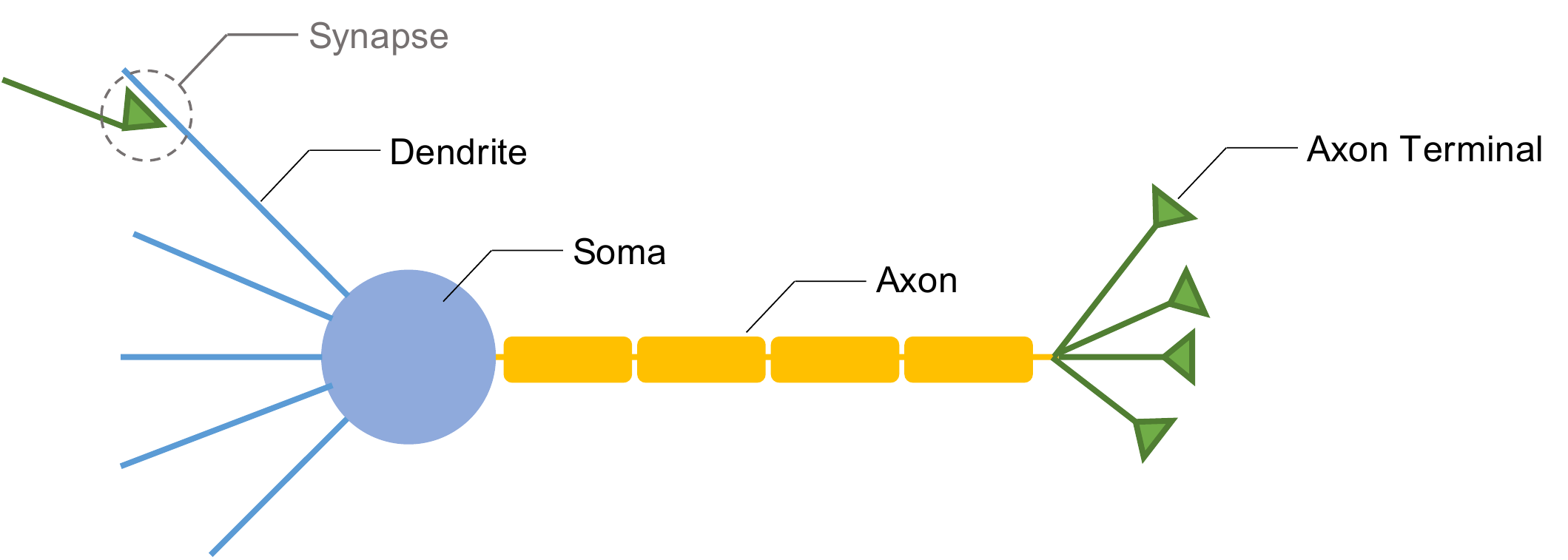} \label{fig:bio_neuron}} \\
        \subfloat[][]{\includegraphics[width=0.8\linewidth]{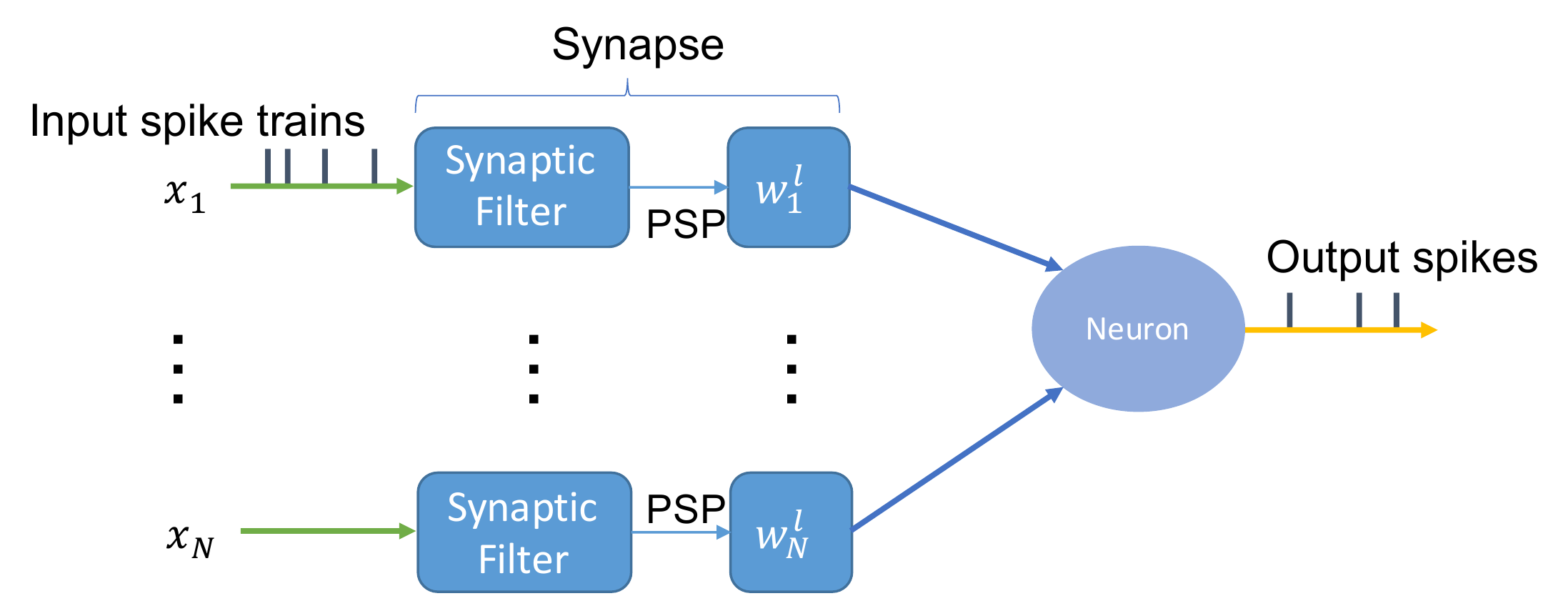} \label{fig:neuron_model}}

        \caption[]{\subref{fig:bio_neuron} Biological-neuron and \subref{fig:neuron_model} spiking-neuron models as infinite impulse response (IIR) filters.}
        \label{fig:neuron}
    \end{figure}

    As shown in Figure \ref{fig:neuron}\subref{fig:bio_neuron}, a biological neuron changes its membrane potential and internal voltage in the soma in response to an input signal from the synapse.
    When the membrane potential exceeds the threshold voltage, a spike signal is generated from the soma and is transmitted to the next neuron through the axon and its terminals.

    As shown in Figure \ref{fig:neuron}\subref{fig:neuron_model}, in many SNN studies, synapses and neurons are treated as separate modules.
    Therefore, synapses receive spike signals from the pre-synaptic neurons and generate a postsynaptic potential (PSP).
    The PSP is weighted for each synapse and flows into the postsynaptic neuron.
    Biologically, this is the input current to the neuron.
    The membrane potential of each neuron varies with the input current (i.e., weighted PSP) over time.
    When this value exceeds the threshold voltage, an output spike is generated from the neuron.
    Immediately afterward, its membrane potential plummets to the resting voltage.
    Because each neuron repeats the above temporal behavior, information flows only via the spike signals, ultimately realizing lower power consumption over time.

    \subsubsection{Neuron and Synapse Model}

    In this study, we use the leaky integrate-and-fire (LIF) neuron model~\citep{steinTHEORETICALANALYSISNEURONAL1965}, which is the most basic and widely used phenomenological neuron model~\citep{schumanSurveyNeuromorphicComputing2017}.
    The synapses are based on the IIR equations~\citep{fang2020exploiting}.
    Because the form in the discrete-time domain, SNNs can be interpreted as a network of IIR filters:

    \begin{subequations}
        \begin{align}
            & V_i^l[t] = \lambda V_i^l[t-1] + I_i^l[t] - V_{th} R_i^l[t]    \label{eq:membrane_potential} \\
            & I_i^l[t] = \sum_j^{N_{l-1}} w_{i,j}^l F_j^l[t] \\
            & R_i^l[t] = \theta R_i^l[t-1] + O_i^l[t-1] \\
            & F_j^l[t] = \beta_{j,0}^l O_j^{l-1}[t] \label{eq:psp} \\
            & O_i^l[t] = U(V_i^l[t] - V_{th}) \\
            & U(x) = \begin{cases}
                0 & \text{if } x < 0 \\
                1 & \text{otherwise}
            \end{cases}, \label{eq:step_func}
        \end{align}
    \end{subequations}

    where $l$ and $i$ denote the index of the layer and the neuron, respectively, $j$ denotes the input index, $t$ is the time step, and $N_l$ is the number of neurons in the $l$-th layer.
    $V_i^l[t]$ is the neuron membrane potential, and $V_{th}$ is the threshold.
    A neuron fires when its membrane potential overcomes it.
    $I_i^l[t]$ is the weighted PSP input.
    $R_i^l[t]$ is the reset voltage used to decrease the membrane potential to the resting voltage after the neuron fires.
    $F_j^l[t]$ is the PSP.
    $O_i^l[t]$ is a spike function that describes the conditions of firing, and $U(x)$ is a Heaviside step function.
    $P$ and $Q$ denote the feedback and feed-forward orders, respectively.
    $\lambda, \theta, \alpha_{j,p}^l$, and $\beta_{j,q}^l$ are the coefficients of the neuron filter, reset filter, and synapse filter, respectively.
    By changing these coefficients, this model can represent various types of neurons.

    \subsubsection{Approximated Back-propagation}   \label{sec:approx_backprop}

    Because this study focuses on image classification tasks, we utilize cross-entropy loss $E$ with the probability of each neuron in the last layer firing in a fixed-length time window, $T$:

    \begin{subequations}
        \begin{align}
            & E = -\sum_i^{N_L} y_i \ln{p_i} \\
            & p_i = \frac{\exp(\sum_t^T O_i^L[t])}{\sum_{j=1}^{N_L} \exp(\sum_t^T O_j^L[t])},
        \end{align}
    \end{subequations}

    where $y_i$ is a one-hot vector representing the truth label, $L$ is the number of layers of the SNN, and $O_i^L[t]$ denotes the output of the last layer.

    Using the chain rule, the gradient can be represented as

    \begin{equation}
        \frac{\partial E}{\partial w^l} = \sum_{t=1}^T \bm{\delta}^l[t] \bm{\epsilon}^l[t] \left( \bm{F}^l[t] + \sum_{i=1}^{t-1} \bm{F}^l[i] \prod_{j=i}^{t-1} \bm{\kappa}^l[j] \right),
    \end{equation}

    where

    \begin{dgroup}
        \begin{dmath}
            \delta^l[t] = \frac{\partial E}{\partial O_i^l[t]} = \sum_{q=0}^Q \sum_j^{N_{l+1}} \frac{\partial E}{\partial O_j^{l+1}[t+q]} \frac{\partial O_j^{l+1}[t+q]}{\partial O_i^l[t]} + \frac{\partial E}{\partial O_i^l[t+1]} \frac{\partial O_i^l[t+1]}{\partial O_i^l[t]}
        \end{dmath}
        \begin{dmath}
            \kappa_i^l[t] = \frac{\partial V_i^l[t+1]}{\partial V_i^l[t]} = \lambda - V_{th} \epsilon_i^l[t]
        \end{dmath}
        \begin{dmath}
            \epsilon_i^l[t] = \frac{\partial U(V_i^l[t] - V_{th})}{\partial V_i^l[t]}.
        \end{dmath}
    \end{dgroup}

    Here, $U(x)$ is not differentiable, which is the main obstacle to SNN learning.
    Fang \etal\ assumed that when Gaussian noise $\mathcal{N}(0,\,\sigma^2)$ was nipped in, LIF neurons could be approximated by Poisson neurons~\citep{fang2020exploiting}, such that the firing rate becomes

    \begin{equation}
        P(v) = \frac{1}{2} \erfc{\frac{V_{th} - v}{\sqrt{2} \sigma}},
    \end{equation}

    where $\erfc{\cdot}$ represents a complementary error function.
    Thus, the derivative of $U(x)$ can be approximated to~\citep{neftci2019surrogate}

    \begin{equation}
        \frac{\partial U(x)}{\partial v} \approx \frac{\partial P(v)}{\partial v} = \frac{e^{-\frac{(V_{th}-v)^2}{2\sigma^2}}}{\sqrt{2 \pi}\sigma}.
    \end{equation}

    \subsection{Information Representation}

    Although the form of the spiking signal varies per neuron type, SNNs are designed based on the assumption that information is represented by spikes.
    Therefore, we must consider how to represent the concerned information in this fashion.

    There are two types of information coding used for this purpose: rate and time.
    Rate coding expresses information in terms of the frequency of the firing of spikes over time.
    However, the time interval between spikes is meaningless.
    Conversely, time coding represents information based on the time interval of each spike.

    Although the time-coding representation is valid~\citep{kumarSpikingActivityPropagation2010,rullenRateCodingTemporal2001}, we apply rate coding for this research, because it is more commonly used in this field to convert continuous-valued input data to event-driven data prior to the first layer in an SNN.

\section{Proposal}

    HNNs have a coding module and two types of neural layers.
    Owing to the format of the data handled by the AL, in which the data are continuous values, and the SL, in which the spike trains exist along the time dimension, the data should be converted from continuous values into spike trains.
    There are two types of ALs that can be incorporated into an SNN: those with fixed weights and those that are trained simultaneously with SLs.

    In the following sections, we detail these two points.

    \subsection{Learning Strategy}

    There are two learning methods regarding ALs and SLs for HNNs: train them separately or simultaneously.

    In the separate method, the ANN is trained in advance.
    However, when building an HNN, the corresponding ALs are removed and connected to the SLs while maintaining the AL weights.
    As shown in Figure \ref{fig:networks}\subref{fig:trainable-als-and-sls-with-duplicate-coding}, when training the HNN, the weights of the extracted ALs are fixed, and only the weights of the SLs are updated.
    In this type of network, because the ALs do not need to learn with the SLs, non-differentiable coding functions can be used to convert the latent vectors from the ALs into spikes.
    Notably, this method requires less time.

    In the simultaneous method, both ALs and SLs are combined from the beginning, and all layers are trained from scratch via back-propagation (Figure \ref{fig:networks}\subref{fig:fixed-als-and-trainable-sls-with-duplicate-coding}).
    Between the last AL and the first SL, coding is required, and this should be differentiable to support back-propagation.
    Therefore, some coding methods are not available from the spatio-temporal back-propagation method (see Section \ref{sec:approx_backprop}).

    \subsection{Rate-based Coding}

    Rate-based coding is utilized at the interface between ALs and SLs, as shown in Figures \ref{fig:networks}\subref{fig:trainable-als-and-sls-with-duplicate-coding} and \ref{fig:networks}\subref{fig:fixed-als-and-trainable-sls-with-duplicate-coding}.

    Network performance in this study is evaluated using continuous-data image input direct to the ALs. Hence, the spiking layers require an additional time dimension. We explore three methods to handle this: duplicate, Gaussian, and Poisson coding.
    This section describes the details.

    \subsubsection{Duplication}    \label{sec:duplicating}

    Expanding the output vector, $\bm{a}$, along the time axis is the most straightforward way to perform rate-based coding.
    The input for the first SL, which is also the coded output of the last AL, is $O_i^{AL}[t] = a_i$.
    Instead of producing a spiking train, this continuous value train is the input for the spiking neurons in the first SL.

    \subsubsection{Gaussian Distribution}  \label{sec:gaussian_dist}

    The duplicating method deterministically codes a continuous value to a continuous-valued train, whereas biological networks are stochastic, owing to uncertainty factors in nature (e.g., noise).
    This method stochastically converts the continuous values generated from the artificial layer output using a reparameterization trick~\citep{kingma2014autoencoding} for connectivity to back-propagation.

    The last artificial layer outputs $\bm{\mu}$ and $\ln{\bm{\sigma}}$ instead of a vector, and the first SL receives re-parameterized continuous-valued trains, $O_i^{AL}[t] \sim \mathcal{N}(\mu, \sigma^2)$ (Figure \ref{fig:networks}\subref{fig:trainable-als-and-sls-with-gaussian-coding}).
    Via the reparameterization trick, the input of the first SL is defined as

    \begin{equation}
        O_i^{AL}[t] = \mu_i + \sigma_i \varepsilon,
    \end{equation}

    where $\varepsilon$ is an auxiliary noise variable, $\varepsilon \sim \mathcal{N}(0, 1)$.

    The parameterization trick makes the boundary between AL and SL differentiable. However, compared with other coding methods, it requires twice the number of neurons in the last AL to adequately measure both performances.
    Thus, although this coding method can theoretically be used in any form of network, SLs cannot be connected with pretrained ALs in this way.

    \subsubsection{Poisson Coding}

    Poisson coding, based on the Poisson-process probability theory, is a widely used coding method~\citep{diehlUnsupervisedLearningDigit2015,querliozImmunityDeviceVariations2013,bingBiologicalinspiredHierarchicalControl2019a} that generates a spike train from a continuous value according to the Poisson distribution, which stochastically converts a continuous value along the time dimension.

    With Poisson coding, the converting function outputs a sequence of spikes such that the time difference between them follows a Poisson distribution, which is not differentiable, unlike the above two methods (Section \ref{sec:duplicating} and \ref{sec:gaussian_dist}).
    Thus, the Poisson coding cannot be used with the simultaneous learning method.

    \subsection{Proposed Models}

    As described, the two types of AL (i.e., fixed and trainable) and the three coding methods (i.e., duplicating, Gaussian, and Poisson) have limited combinations, owing to their traits. In this study, we investigate the possibility of HNNs that use the four methods described in the subsequent sections.
    For simplicity, we assume that all networks in this study have a feed-forward architecture.

\begin{figure}[htbp]
        \centering
        \subfloat[][Trainable AL and SL network with duplicate coding]{\includegraphics[width=0.60\linewidth]{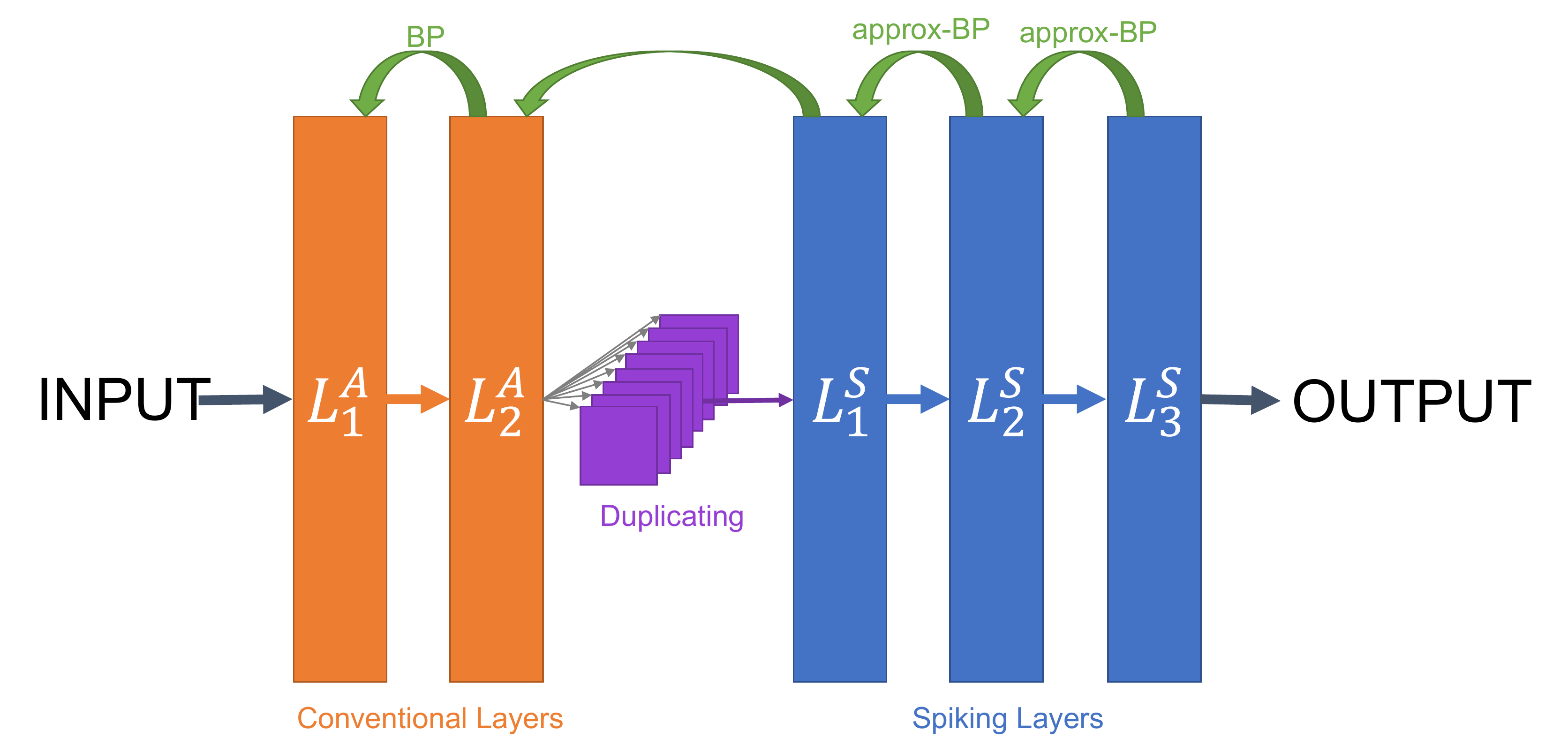} \label{fig:trainable-als-and-sls-with-duplicate-coding}} \\
        \subfloat[][Fixed AL and Trainable SL  network with duplicate coding]{\includegraphics[width=0.60\linewidth]{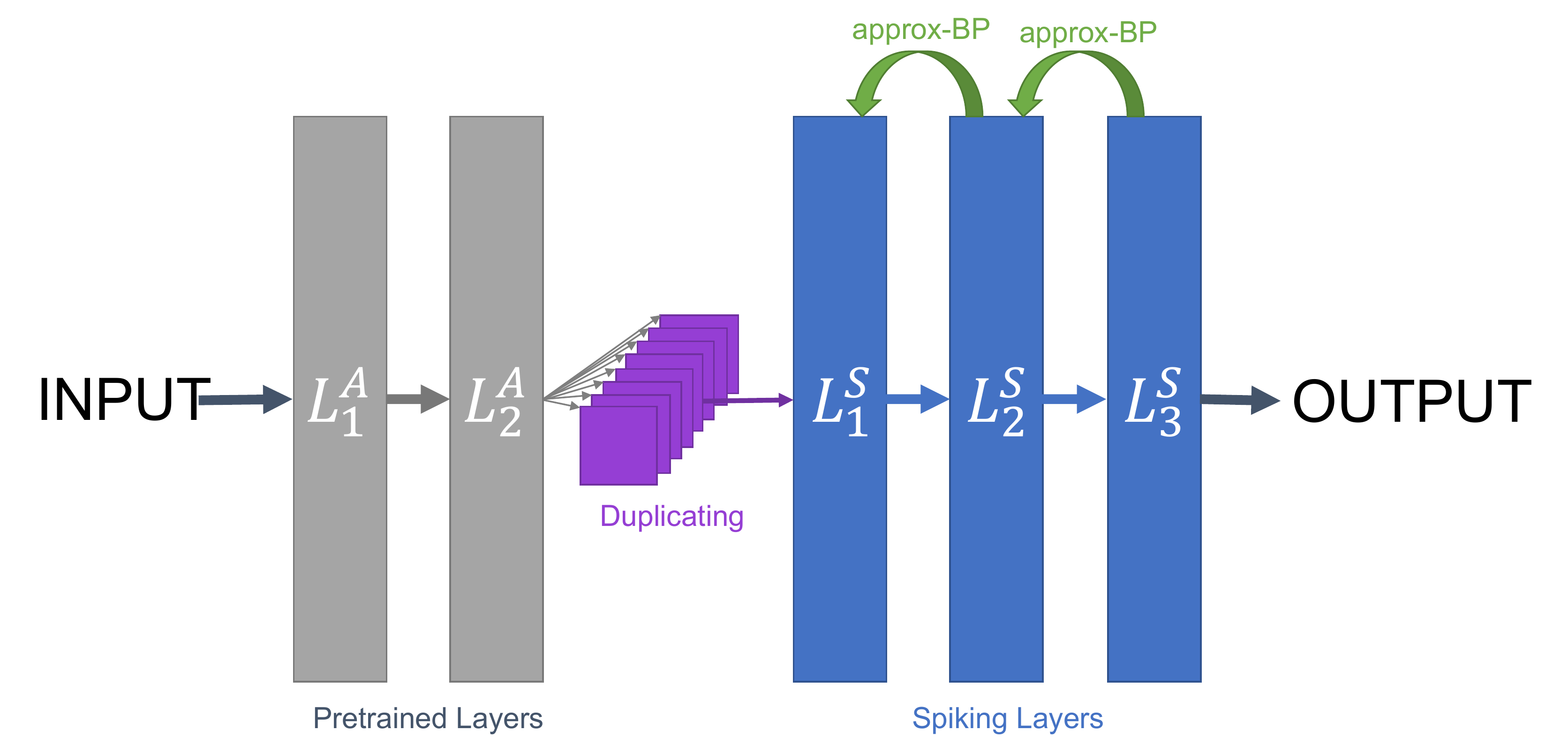} \label{fig:fixed-als-and-trainable-sls-with-duplicate-coding}} \\
        \subfloat[][Trainable AL and SL network with Gaussian coding]{\includegraphics[width=0.60\linewidth]{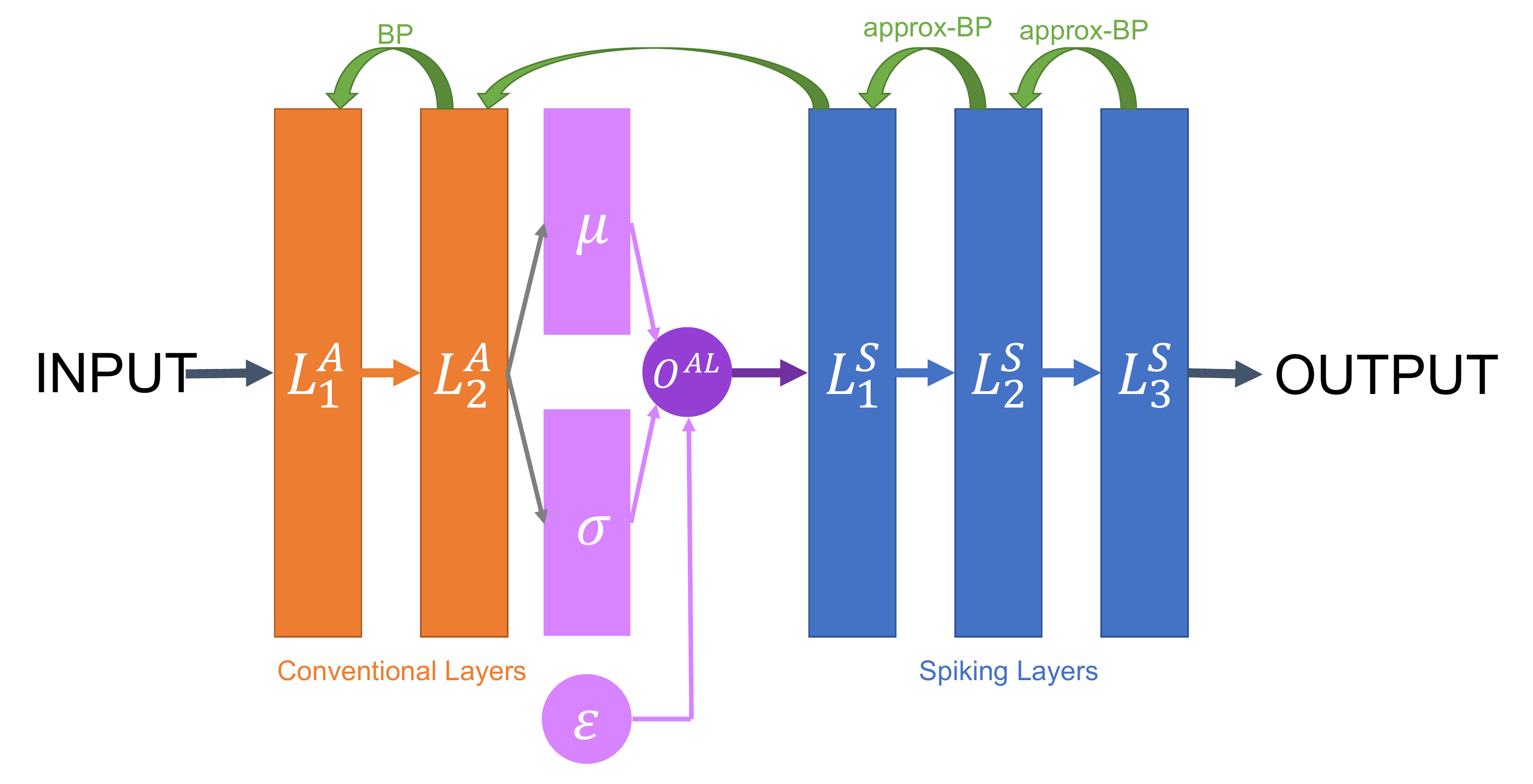} \label{fig:trainable-als-and-sls-with-gaussian-coding}} \\
        \subfloat[][Fixed AL and Trainable SL network with Poisson coding]{\includegraphics[width=0.60\linewidth]{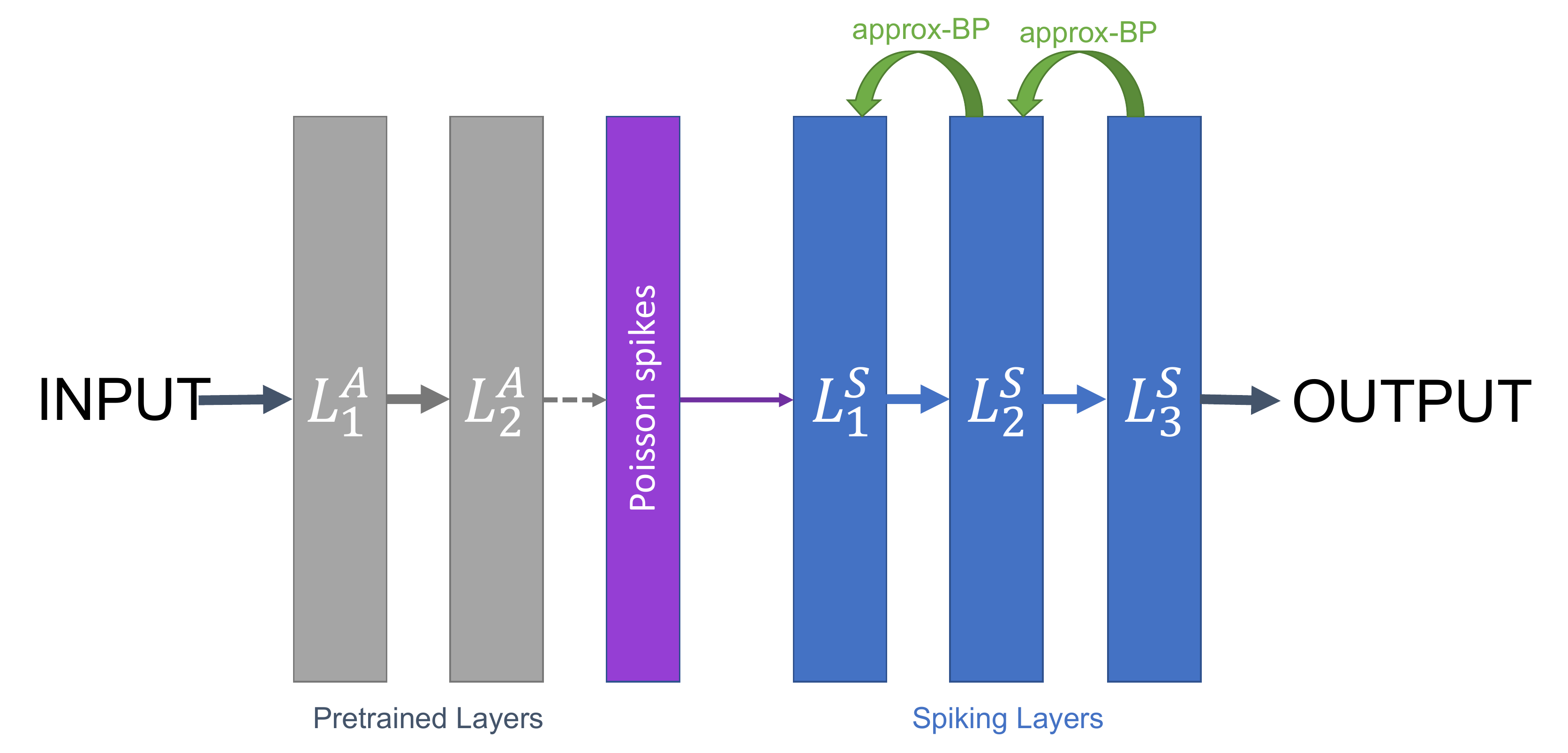} \label{fig:fixed-als-and-trainable-sls-with-poisson-coding}} \\

        \caption[]{Hybrid Neural Networks.}
        \label{fig:networks}
    \end{figure}

    Figures \ref{fig:networks}\subref{fig:trainable-als-and-sls-with-duplicate-coding}--\subref{fig:fixed-als-and-trainable-sls-with-poisson-coding} show all of the neural networks implemented in this paper.

    The main limiting factor in developing an HNN is determining whether or not the coding method is differentiable.
    When fixed or pretrained ALs are incorporated into the HNN, all coding methods can be adopted.
    Conversely, when a trainable AL is used, only differentiable methods can be adopted.

    Moreover, for evaluation, the Gaussian coding implemented by the reparameterization trick cannot be combined with HNNs having pretrained ALs, because the reparameterization trick requires more neurons than are available in the basic network architecture.

    Trainable ALs and SLs in a duplicate coding network are shown in Figure \ref{fig:networks}\subref{fig:trainable-als-and-sls-with-duplicate-coding}. Both the ALs and the SLs are trained simultaneously via back-propagation using the duplicating method for coding after the last AL.

    Fixed ALs and trainable SLs within a duplicating coding network (Figure \ref{fig:networks}\subref{fig:fixed-als-and-trainable-sls-with-duplicate-coding}) optimize the weights only for SLs during training. To facilitate the addition of the time dimension between the AL and the SL, the network merely copies the output from the last AL over a certain number of time steps.

    As shown in Figure \ref{fig:networks}\subref{fig:trainable-als-and-sls-with-gaussian-coding}, the trainable ALs and SLs within the Gaussian-coding network are trained for all layers together, and, even with stochastic coding, the training phase follows a Gaussian distribution.
    This network takes the longest time to train, because, unlike networks having fixed ALs, all layers should be trained, and the reparameterization trick used with Gaussian coding becomes more computationally complex than the straightforward duplication.

    With fixed ALs and trainable SLs within a Poisson coding network (Figure \ref{fig:networks}\subref{fig:fixed-als-and-trainable-sls-with-poisson-coding}), Poisson coding is used to generate spike trains from the latent vector rendered from the pretrained ALs.
    Owing to the undifferentiable coding method, only the SLs are trained.

\section{Experiments}
    \subsection{Dataset}

    In this study, two datasets (i.e., MNIST\citep{LeCun1998} and CIFAR-10\citep{krizhevsky2009learning}) were used to measure network performance.
    Each dataset was divided into three subsets for training, validation, and testing at a ratio of $6:2:2$, respectively.

    Each network was trained with the training dataset for $100$ and $150$ epochs on MNIST and CIFAR-10, respectively.
    At the end of each epoch, network performance was evaluated using the validation dataset, and the network achieving the best performance on the validation dataset was used to compare performance based on the testing dataset.

    \subsection{Receiving Continuous Values}

    There are two ways in which a continuous value train can be inserted into a spiking neuron: as a spike train through the axon or directly.

    The first was used in~\citep{fang2020exploiting}, utilizing back-propagation.
    The continuous-valued input is converted to a PSP through the axon and is treated as a spike train. The spiking neurons in the first layer receives the PSP.

    In the second way, spiking neurons in the first layer directly receives continuous input values instead of the PSP.
    This is more biologically plausible, because the sensory-nerve endings in animals are sensory neurons rather than axons.

    Using MNIST with a multi-layer perceptron (MLP)(\texttt{S784-\allowbreak S500-\allowbreak S10}), the axon-input network achieved $\SI{97.56}{\%}$, and the direct-input network achieved $\SI{97.71}{\%}$.

    The results reveal that the axon-inputting method does not improve network performance.
    Therefore, we used the direct-inputting method in the following experiments.

    \subsection{Hybrid Neural Networks}

    HNNs can be regulated with rate-based coding (e.g., duplication, Poisson, and Gaussian) and trainable layers, where the SLs are combined with fixed or trainable ALs.
    Notably, the fixed-AL and trainable-SL network with Gaussian coding is not compared here, because it relies on the reparameterization trick and different architecture.

    In the network having a fixed AL and a trainable SL, the fixed ALs were generated from a pure ANN, which has the highest accuracy when trained on the training dataset and evaluated on the validation dataset.

    In Tables \ref{tab:duplicating_result}\subref{tab:mlp_duplicating}, \ref{tab:gaussian_result}\subref{tab:cnn_duplicating}, \ref{tab:gaussian_result}\subref{tab:mlp_gaussian}, \ref{tab:duplicating_result}\subref{tab:cnn_gaussian}, \ref{tab:poisson_result}\subref{tab:mlp_poisson}, and \ref{tab:poisson_result}\subref{tab:cnn_poisson}, the network architectures are shown using a notation in which \texttt{A}$n$ and \texttt{S}$m$ indicate a layer having $n$ artificial neurons and $m$ spiking neurons, respectively.
    Additionally, \texttt{A}$n$\texttt{C}$k$ or \texttt{S}$n$\texttt{C}$k$ indicate an spiking convolutional AL with kernel $k\times k$.
    Here, the coding method is implicitly inserted between \texttt{A$n$-S$m$}.

    A three-layer MLP having ALs, SLs, and a coding layer in the interface between the AL and SL was used to classify images on the MNIST dataset to consider the effect of combining ALs and SLs.

    As the baseline, a pure SNN using a dual exponential post-synaptic potential kernel with the \texttt{S784-S500-S10} architecture was trained.
    Moreover, the last row in Table \ref{tab:duplicating_result}\subref{tab:mlp_duplicating} shows the result of the pure ANN fully comprising ALs.
    Here, the respective results of the pure SNN and ANN with separate learning cycles are not described in all tables, because they do not need to train each layer separately.

    As shown in Table \ref{tab:duplicating_result}\subref{tab:mlp_duplicating}, the network having fixed ALs and trainable SLs using duplicating coding achieved the highest score.
    Notably, the \texttt{A784-A500-S10} architecture recorded higher performance than did the ANN, although both networks shared weights in the first and second layers.

    \subsubsection{Separate vs. Simultaneous Training}  \label{sec:duplicating_exp}

    To compare separate and simultaneous learning for classification accuracy, we conducted comparison experiments using the above networks with duplicate coding, which can be used for both types of learning.

    Table \ref{tab:duplicating_result}\subref{tab:mlp_duplicating} and \ref{tab:duplicating_result}\subref{tab:cnn_duplicating} show the results of MLPs and convolutional neural networks (CNNs), each having the same number of neurons.
    However, the ratio of ALs to SLs was different.
    In each table, the accuracy of the pure SNN and ANN with fixed ALs is not listed, because neither had pre-training layers.
    Notably, the results of the pure SNN and ANN are similar to those in Table \ref{tab:gaussian_result}, because they used the same architectures with duplicate coding.

    The results in both tables show that, generally, using pure networks achieves higher accuracy.
    Additionally, in HNNs, increasing the percentage of ALs improves performance, revealing a trade-off between computational accuracy and energy efficiency.
    With rate-based coding, an artificial neuron can represent more accurate information.
    However, a spiking neuron can reduce power consumption.
    Furthermore, although the separately trained networks showed higher accuracy than those with simultaneous learning, the performance of the more complex tasks (CIFAR-10) was reduced when the ratio of SLs in the trainable network was too small.
    Thus, for complex tasks in huge networks, training ALs and SLs at the same time gives better accuracy than using only fixed-ALs in a pure SNN.

    \begin{table*}[htbp]
        \centering
        \subfloat[][]{\begin{tabular}{lcc}
    \hline
    & \multicolumn{2}{c}{Traning method} \\
    & Separate & Simultaneous \\
    \hline
    \texttt{S784-S500-S10} & - & $\SI{98.13}{\%}$ \\ 
    \texttt{A784-S500-S10} & $\SI{98.22}{\%}$ & $\SI{97.78}{\%}$ \\   
    \texttt{A784-A500-S10} & $\SI{99.17}{\%}$ & $\SI{98.07}{\%}$ \\   
    \texttt{A784-A500-A10} & - & $\SI{98.23}{\%}$ \\ 
    \hline
\end{tabular}%
 \label{tab:mlp_duplicating}} \qquad
        \subfloat[][]{\begin{tabular}{lcc}
    \hline
    & \multicolumn{2}{c}{Traning method} \\
    & Separate & Simultaneous \\
    \hline
    \texttt{S32C3-S32C3-S64C3-P2-S64C3-P2-S512-S10} & - & $\SI{59.96}{\%}$ \\ 
    \texttt{A32C3-S32C3-S64C3-P2-S64C3-P2-S512-S10} & $\SI{56.01}{\%}$ & $\SI{53.23}{\%}$ \\  
    \texttt{A32C3-A32C3-A64C3-P2-S64C3-P2-S512-S10} & $\SI{63.30}{\%}$ & $\SI{55.22}{\%}$ \\  
    \texttt{A32C3-A32C3-A64C3-P2-A64C3-P2-S512-S10} & $\SI{62.78}{\%}$ & $\SI{66.08}{\%}$ \\  
    \texttt{A32C3-A32C3-A64C3-P2-A64C3-P2-A512-A10} & - & $\SI{70.09}{\%}$ \\ 
    \hline
\end{tabular}%
 \label{tab:cnn_duplicating}}
        \captionsetup{margin=50pt}
        \caption[]{\subref{tab:mlp_duplicating} MLPs with duplicate coding for MNIST and \subref{tab:cnn_duplicating} CNNs with duplicate coding for CIFAR-10}
        \label{tab:duplicating_result}
    \end{table*}

    \subsubsection{Gaussian Coding}

    The results of MLPs and CNNs for MNIST and CIFAR-10, respectively, with Gaussian coding are shown in Tables \ref{tab:gaussian_result}\subref{tab:mlp_gaussian} and \ref{tab:gaussian_result}\subref{tab:cnn_gaussian}.
    As mentioned, Gaussian coding can only be employed with trainable ALs, owing to the additional neurons needed for reparameterization.

    Here, the \texttt{S784-S500-S10} architecture (Table \ref{tab:gaussian_result}\subref{tab:mlp_gaussian}) and \texttt{S32C3-\allowbreak S32C3-\allowbreak S64C3-\allowbreak P2-\allowbreak S64C3-\allowbreak P2-\allowbreak S512-\allowbreak S10} network (Table \ref{tab:gaussian_result}\subref{tab:cnn_gaussian}) coded the input data into spiking trains using the duplicating method, because it was not possible to use Gaussian coding without varying the network architecture. The accuracies of both networks compared with the baseline are similar to those shown in Tables \ref{tab:duplicating_result}\subref{tab:mlp_duplicating} and \ref{tab:duplicating_result}\subref{tab:cnn_duplicating}.
    Moreover, the \texttt{A784-A500-A10} architecture (Table \ref{tab:gaussian_result}\subref{tab:mlp_gaussian}) and \texttt{A32C3-\allowbreak A32C3-\allowbreak A64C3-\allowbreak P2-\allowbreak A64C3-\allowbreak P2-\allowbreak A512-\allowbreak A10} network (Table \ref{tab:gaussian_result}\subref{tab:cnn_gaussian}), listed for only comparison, do not have the coding layer; these networks were directly inferred from the original data, and the results in both tables are the same as those in Tables \ref{tab:duplicating_result}\subref{tab:mlp_duplicating} and \ref{tab:duplicating_result}\subref{tab:cnn_duplicating}.

    From the results (Figures \ref{tab:duplicating_result}\subref{tab:cnn_duplicating} and \ref{tab:gaussian_result}\subref{tab:mlp_gaussian}), except for the \texttt{A32C3-\allowbreak A32C3-\allowbreak A64C3-\allowbreak P2-\allowbreak S64C3-\allowbreak P2-\allowbreak S512-\allowbreak S10} network, the trend of improving accuracy with an increasing AL ratio was maintained using duplicate coding (Section \ref{sec:duplicating_exp}).
    Additionally, compared with the model with trainable ALs and SLs and duplicate coding (Tables \ref{tab:duplicating_result}\subref{tab:mlp_duplicating} and \ref{tab:duplicating_result}\subref{tab:cnn_duplicating}), the accuracies of the models with Gaussian coding (Tables \ref{tab:gaussian_result}\subref{tab:mlp_gaussian} and \ref{tab:duplicating_result}\subref{tab:cnn_duplicating}) were higher for lower AL ratios.

    \begin{table*}[htbp]
        \centering
        \subfloat[][]{\begin{tabular}{lc}
    \hline
    Network architecture & Accuracy \\
    \hline
    \texttt{S784-S500-S10} (baseline) & $\SI{98.13}{\%}$ \\
    \texttt{A784-S500-S10} & $\SI{98.02}{\%}$ \\   
    \texttt{A784-A500-S10} & $\SI{98.03}{\%}$ \\   
    \texttt{A784-A500-A10} & $\SI{98.23}{\%}$ \\
    \hline
\end{tabular}%
 \label{tab:mlp_gaussian}} \qquad
        \subfloat[][]{\begin{tabular}{lc}
    \hline
    Network architecture & Accuracy \\
    \hline
    \texttt{S32C3-S32C3-S64C3-P2-S64C3-P2-S512-S10} (baseline) & $\SI{59.96}{\%}$ \\
    \texttt{A32C3-S32C3-S64C3-P2-S64C3-P2-S512-S10} & $\SI{59.35}{\%}$  \\ 
    \texttt{A32C3-A32C3-A64C3-P2-S64C3-P2-S512-S10} & $\SI{65.30}{\%}$  \\ 
    \texttt{A32C3-A32C3-A64C3-P2-A64C3-P2-S512-S10} & $\SI{63.82}{\%}$  \\    
    \texttt{A32C3-A32C3-A64C3-P2-A64C3-P2-A512-A10} & $\SI{70.09}{\%}$ \\
    \hline
\end{tabular}%
 \label{tab:cnn_gaussian}}
        \captionsetup{margin=50pt}
        \caption[]{\subref{tab:mlp_gaussian} MLPs with Gaussian coding for MNIST and \subref{tab:cnn_gaussian} CNNs with Gaussian coding for CIFAR-10}
        \label{tab:gaussian_result}
    \end{table*}

    \subsubsection{Poisson Coding}

    Poisson coding is a mathematical technique used to generate spike trains that follow a Poisson distribution from continuous values.
    The Poisson process is usually not differentiable. Thus, trainable ALs and SLs cannot be used.
    To apply Poisson coding, the continuous values should be normalized to $(0,1)$.
    In this experiment, for simplification, we used the sigmoid function. The output values are $(0,1)$ for the activation function instead of the rectified linear unit, which are utilized in the pretrained model with ALs.

    As shown in Tables \ref{tab:poisson_result}\subref{tab:mlp_poisson} and \ref{tab:poisson_result}\subref{tab:cnn_poisson}, using Poisson coding instead of duplicate or Gaussian coding dramatically decreased the performance of every network architecture.
    Unlike previous results, the continuous-valued data were converted into spike trains with Poisson coding for pure SNNs, the \texttt{S784-S500-S10} architecture (Table \ref{tab:poisson_result}\subref{tab:mlp_poisson}), and the \texttt{S32C3-S32C3-S64C3-P2-S64C3-P2-S512-S10} network (Table \ref{tab:poisson_result}\subref{tab:cnn_poisson}).
    The last row of each table shows the result of the pretrained ANN, which is the basis of ALs in HNNs.

    From the results in Table \ref{tab:poisson_result}\subref{tab:mlp_poisson}, even such low accuracies follow this tendency: ALs lead to higher accuracy.
    In contrast, the results in Table \ref{tab:poisson_result}\subref{tab:cnn_poisson} show the opposite, and it is observed that Poisson coding generates too much noise, leading to a decrease in accuracy rather than improving the generalization.
    Notably, these noisy latent vectors reduce the accuracy of image recognition.

    \begin{table*}[htbp]
        \centering
        \subfloat[][]{\begin{tabular}{lc}
    \hline
    Network architecture & Accuracy \\
    \hline
    \texttt{S784-S500-S10} (baseline) & $\SI{98.09}{\%}$ \\ 
    \texttt{A784-S500-S10} & $\SI{89.67}{\%}$ \\    
    \texttt{A784-A500-S10} & $\SI{95.93}{\%}$ \\    
    \texttt{A784-A500-A10} (pretrained) & $\SI{97.67}{\%}$ \\   
    \hline
\end{tabular}%
 \label{tab:mlp_poisson}} \qquad
        \subfloat[][]{\begin{tabular}{lc}
    \hline
    Network architecture & Accuracy \\
    \hline
    \texttt{S32C3-S32C3-S64C3-P2-S64C3-P2-S512-S10} (baseline) & $\SI{59.83}{\%}$ \\    
    \texttt{A32C3-S32C3-S64C3-P2-S64C3-P2-S512-S10} & $\SI{50.28}{\%}$ \\  
    \texttt{A32C3-A32C3-A64C3-P2-S64C3-P2-S512-S10} & $\SI{36.94}{\%}$ \\  
    \texttt{A32C3-A32C3-A64C3-P2-A64C3-P2-S512-S10} & $\SI{31.32}{\%}$ \\  
    \texttt{A32C3-A32C3-A64C3-P2-A64C3-P2-A512-A10} (pretrained) & $\SI{56.43}{\%}$ \\   
    \hline
\end{tabular}%
 \label{tab:cnn_poisson}}
        \captionsetup{margin=50pt}
        \caption[]{\subref{tab:mlp_poisson} MLPs with Poisson coding for MNIST and \subref{tab:cnn_poisson} CNNs with Poisson coding for CIFAR-10}
        \label{tab:poisson_result}
    \end{table*}

\section{Discussion}

    The results (Figures \ref{tab:duplicating_result}\subref{tab:mlp_duplicating}--\subref{tab:cnn_poisson}) largely delineate a tendency of improving accuracy with the increase in the percentage of ALs in the hybrid model.
    This was true, except for CNNs having two ALs and three SLs (i.e., \texttt{A32C3-A32C3-A64C3-P2-S64C3-P2-S512-S10}).
    This confirms our expectation that ANNs are better suited to handle continuous values.

    The highest scores of the HNNs that used Gaussian coding were not greater than those using duplicate coding.
    However, networks having fewer ALs (Tables \ref{tab:gaussian_result}\subref{tab:mlp_gaussian} and \ref{tab:gaussian_result}\subref{tab:cnn_gaussian}) performed better than networks with trainable ALs and SLs using duplicate coding, especially for CIFAR-10.
    Therefore, although Gaussian coding did not show a clear advantage in the experiments, high performance can be expected in more complex tasks in the future.

    In these experiments, we did not find any advantage from using Poisson coding.
    In fact, when used with an HNN, its accuracy was greatly reduced.
    This is probably because there was too much information missing, owing to the noise generated during the coding process.
    In contrast, pure SNNs showed competitive accuracy with other coding methods, as demonstrated in previous studies~\citep{Deng2020}.
    Interestingly, for HNNs having pretrained ALs (Figure \ref{tab:duplicating_result}\subref{tab:cnn_poisson}), the CNNs having fewer ALs outperformed those with more.
    This suggests that the use of a spiking conventional layer immediately after coding contributes to performance improvement, and this inclination is not seen in the MLP results.

\section{Conclusion}

    In this study, we proposed hybrid neural networks that combined features of the conventional continuous-input ANN with those of the more bio-plausible, event-driven SNN.
    The elements comprising the HNN, including the various ALs that were combined with the SNN features and respective coding methods, were evaluated on image classification tasks using MNIST and CIFAR-10 datasets.
    Several HNNs exceeded the performance of pure SNNs and demonstrated notable effectiveness.
    Moreover, it was shown that the performance changes depended on the ratio of ALs and SLs and the coding method used to transform the continuous values to spiking trains for connecting the ALs to the SLs.
    Additionally, even when the most straightforward duplicate coding method was used, the performance was equal to or better than the other coding methods.

    We believe that our work helps further increase the potential of SNNs, which are more energy-efficient than ANNs but are disadvantageous for handling continuous-valued data input.

\bibliographystyle{unsrtnat}
\bibliography{references}  






\end{document}